\documentclass[a4paper]{article}

\usepackage{arxiv}

\usepackage[utf8]{inputenc} 
\usepackage[T1]{fontenc}    
\usepackage[breaklinks]{hyperref}
\usepackage{subfigure}
\usepackage{url}
\usepackage{breakurl}
\usepackage{booktabs}       
\usepackage{amsfonts}       
\usepackage{nicefrac}       
\usepackage{microtype}      
\usepackage{amsmath}
\usepackage{amsfonts}
\usepackage{cite}
\usepackage{amssymb}
\usepackage{amsthm}
\usepackage{verbatim}
\usepackage{enumerate}
\usepackage{mdwlist} 
\usepackage{color}
\usepackage[utf8]{inputenc}

\usepackage{algpseudocode}
\usepackage{algorithmicx}
\usepackage{lineno}
\usepackage{framed,multirow}
\usepackage[final]{pdfpages}
\usepackage{latexsym}
\usepackage{wrapfig}

\usepackage{verbatim}
\usepackage{bm}
\usepackage{authblk}

\expandafter\def\expandafter\UrlBreaks\expandafter{\UrlBreaks
    \do\a\do\b\do\c\do\d\do\e\do\f\do\g\do\h\do\i\do\j%
    \do\k\do\l\do\m\do\n\do\o\do\p\do\q\do\r\do\s\do\t%
    \do\u\do\v\do\w\do\x\do\y\do\z\do\A\do\B\do\C\do\D%
    \do\E\do\F\do\G\do\H\do\I\do\J\do\K\do\L\do\M\do\N%
    \do\O\do\P\do\Q\do\R\do\S\do\T\do\U\do\V\do\W\do\X%
    \do\Y\do\Z\do\/\do-}

\title{Fusion of complex networks and randomized neural networks for texture analysis}

\author[1]{Lucas C. Ribas}
\author[3]{Jarbas Joaci de Mesquita S\'{a} Junior}
\author[2]{Leonardo F. S. Scabini}
\author[1,2]{Odemir M. Bruno}
\affil[1]{\small{S\~{a}o Carlos Institute of Physics, University of S\~{a}o Paulo (USP), PO Box 369, 13560-970, S\~{a}o Carlos, SP, Brazil. \protect\\Scientific Computing Group}}
\affil[2]{\small{Institute of Mathematics and Computer Science, University of S\~{a}o Paulo (USP), USP, Avenida Trabalhador s\~ao-carlense, 400, 13566-590, S\~ao Carlos, SP, Brazil.}}
\affil[3]{\small{Curso de Engenharia da Computa\c{c}\~ao,
Programa de P\'os-Grad. em Eng. El\'etrica e de Computa\c{c}\~ao, 
Campus de Sobral, Universidade Federal do Cear\'a,
Rua Coronel Estanislau Frota, 563, Centro,
Sobral, Cear\'a, CEP: 62010-560, Brasil}}

\begin{document}
\maketitle

\begin{abstract}
This paper presents a high discriminative texture analysis method based on the fusion of complex networks and randomized neural networks. In this approach, the input image is modeled as a complex networks and its topological properties as well as the image pixels are used to train randomized neural networks in order to create a signature that represents the deep characteristics of the texture. The results obtained surpassed the accuracy of many methods available in the literature. This performance demonstrates that our proposed approach opens a promising source of research, which consists of exploring the synergy of neural networks and complex networks in the texture analysis field.
\end{abstract}


\section{Introduction}
\label{sec:introduction}
Most of computer vision applications consider texture as a key factor to image discrimination, thus texture analysis has been a constant research field since the 1960s. The texture is a visual pattern related to the object surface, which in an image is represented by the pixel spatial organization. However, the interpretation of texture is ambiguous, thus there is no formal definition to the term that is widely accepted by the scientific community. This resulted in an extensive and heterogeneous literature of texture analysis methods proposed along the years \cite{zhang2002survey,liu2018texturesurvey,BHARATI200457}. Usually, texture descriptors are applied in different areas such as industrial inspection \cite{industrial2003}, geology \cite{2013geology}, medicine \cite{2013tumor}, material science \cite{zimer2011investigation}, and so on.
    
Classical texture analysis techniques can be grouped into four different approaches: statistical, spectral, structural, model-based methods \cite{methods}. The earlier and most diffused methods are statistical-based, such as variants of gray-level co-occurrence matrices (GLCM) \cite{haralick1973, PALM2004integrativeCooccurrence} and local binary patterns (LBP) \cite{nanni2012survey,brahnam2016local}. Spectral methods explore texture in the frequency domain, some examples are Gabor filters \cite{1990gabor} and wavelet transform \cite{DEVES20142925}.
On the other hand,  structural methods consider texture as a combination of  smaller elements, called textons, that compose the overall texture as a spatially organized pattern. 
A common approach of this kind of analysis is the Morphological decomposition \cite{lam1997rotated}. 
Finally, model-based methods represent textures through sophisticated mathematical models and estimating its parameters. Common methods of this category include Fractal models \cite{backes2012,backes2009plant,backes2008fractal,dalcimar,Ribas2015,BrunoPFC08} and stochastic models \cite{panjwani1995markov}

Besides classical methods, more recent and innovative techniques are addressing texture differently, achieving promising results. An example is the set of techniques that use learning, such as descriptors based on a vocabulary of scale invariant feature transform (SIFT) \cite{2004densesift}, often called bag-of-visual-words (BOVW). Methods based on image complexity analysis are also gaining attention such as cellular automata \cite{CA2015texture} and complex networks (CN) \cite{backes2012,scabini2015texture,xu2015complex,gonccalves2016texture,scabini2019multilayer}. In particular, methods based on the CN theory have achieved promising results due to its capacity to represent the relation among structural elements of texture.
However, the problem of how to achieve more satisfactory modeling (i.e., a lesser number of parameters) and new ways of characterizing the network remains a challenge to overcome.

In this paper, we propose a novel approach that combines complex networks and randomized neural networks (RNN) in order to obtain a texture signature. Complex networks is attracting increasing attention due to its flexibility and generality for representing many real-world systems, including texture images. On the other hand, a randomized neural network is a neural network with a unique hidden layer and a very fast learning algorithm, which has been used in many pattern recognition tasks.
Here we first model the texture image as a directed network, representing the information about the pixels and its neighbors as vertices and edges. To characterize the texture, the topological properties from the modeled network and the image pixels are used to train a randomized neural network, and the set of output weights is used as a feature vector that represents discriminative characteristics of the texture. Experimental results on four databases demonstrated a better performance of the proposed method when compared to other methods of the literature.

The remainder of this paper is organized as follows. Section \ref{sec:back} describes the fundamentals of complex networks and randomized neural networks. A novel method for texture classification based on fusion of complex networks and randomized neural networks is presented in Section \ref{sec:propose}. Section \ref{sec:experiments} describes the databases and experiments performed to evaluate the proposed method. The discussion about the results achieved and comparisons are presented in Section \ref{sec:results}. Finally, in Section \ref{sec:conclusion}, we conclude the work with some remarks.

\section{Background} \label{sec:back}

\subsection{Complex networks}

Almost any natural phenomena can be modeled as networks by defining a set of entities and establishing a criterion of relation between them. 
Some classical examples are the internet, composed of various connected computers and routers, and a network of a cell, describing chemicals connected by chemical reactions.
Complex networks are part of an area known as network science \cite{barabasi2016network}. Network science is strongly based on graph theory. In the last decades, works have shown patterns present in many networks or graphs, which were then understood as a structural characteristic of some models such as the scale-free \cite{scalefreeCN} and the small-world \cite{smallworldCN}. These findings have caused increasing interest from the scientific community on the study of complex networks, creating a new multidisciplinary research field. 

The theoretical foundations of this area arise from of the intersection of the graph theory, physics, mathematics, statistics and computer science.  
Therefore, CN has been employed as a powerful tool for pattern recognition \cite{Miranda2016}, where natural systems of many areas are modeled as networks and then quantified through its topological structure. 
CN applications are found in various areas of science, such as, physics, social sciences, biology, mathematics, ecology, medicine, computer science, linguistic, neuroscience, among others \cite{cnAppSurvey2011}.

Formally, a network or graph $G$ is described by a tuple of vertices and edges $(V, E)$. Let $v_i$ be a vertex of the set $V=\{v_1,...,v_n\}$. An edge $e_{v_i,v_j}$ represents a connection between two vertices $v_i$ and $v_j$, so the set $E=\{e_{v_i,v_j},...\}$ is composed of all edges connecting vertices of $V$. 
The network can also be directed, in this case, the edges $e_{v_i,v_j}$ have a direction from $v_i$ to $v_j$.
In most of the CN applications, the first step is to define how to model the target problem as a network, thus defining what are the vertices and what are the edges. 
Once $G$ is properly built, many measures can be computed to quantify its structure, varying from centrality, path-based measures, community structure, and many more \cite{costa2007CNsurvey}. 
Moreover, the structure of a real network is the result of the continuous evolution of the forces that formed it, and certainly affects the function of the system \cite{structureanddynamics}. 
Therefore, the network dynamics can be analyzed by the characterization of its structural evolution in function of time or some modeling parameter.

\subsection{Randomized neural networks}

Randomized neural networks \cite{schmidt1992feedforward,pao1992functional,pao1994learning,huang2006extreme} are neural networks composed of two neuron layers (hidden and output layer), each one with a different role in the regression/classification task. The hidden layer has its neural weights determined randomly according to a probability distribution (for instance, a uniform or normal distribution). Its purpose is to project non-linearly the input data in another dimensional space where it is more likely that the feature vectors are linearly separable, as stated in Cover's theorem \cite{Cover1965}. In turn, the output layer aims to linearly separate these projected feature vectors using the least-squares method.

Mathematically, letting $X=[\vec{x_1},\vec{x_2},\ldots,\vec{x_N}]$ be a matrix of input feature vectors (including $+1$ for bias weight) and $D=[\vec{d_1},\vec{d_2},\ldots,\vec{d_N}]$ be the corresponding labels, the first step is to build the matrix of hidden neuron weights $W$ of dimensions $Q \times (p+1)$, where $Q$ and $p$ are the number of hidden neurons and the number of attributes in each input feature vector, respectively.  

Next, the output of the hidden layer for all the feature vectors $\vec{x_i}$ $(i \in {1,\ldots,N})$ can be obtained by $Z=\phi(WX)$, where $\phi(.)$ is generally a sigmoid or hyperbolic tangent function. Thus, this matrix of projected vectors $Z=[\vec{z_1},\vec{z_2},\ldots,\vec{z_N}]$ (including $+1$ for bias weight) can be used to compute the output neuron weights, according to the following equation

\begin{equation}
M = DZ^T(ZZ^T)^{-1},
\label{eq:rnn}
\end{equation}
where $Z^T(ZZ^T)^{-1}$ is the Moore-Penrose pseudo-inverse \cite{Moore1920,penrose_1955}. 

Sometimes, the matrix $ZZ^T$ becomes singular (that is, without inverse), or close to singular, which results in unstable results in Equation \ref{eq:rnn}. In order to avoid these drawbacks, it is possible to use the Tikhonov regularization \cite{tiknonov1963,calvetti2000}, according to

\begin{equation}
M = DZ^T(ZZ^T+\lambda I)^{-1},
\end{equation}
where $0<\lambda<1$ and $I$ is the identity matrix.

\section{Proposed method} \label{sec:propose}

In this section, we describe the proposed method that combines a new texture modeling in complex networks and randomized neural networks for texture characterization.

\subsection{Modeling texture as directed CN}

Let $I$ be an image composed of pixels $i$, which have as Cartesian coordinates $x_i$ and $y_i$. In gray-scale images, each pixel has an intensity represented by an integer value $I(i) \in [0,L]$, where $L$ is the highest gray-level value. To model a texture image as a directed network, each pixel $i$ is mapped as a vertex $v_i \in V$ of a network $R$. The set of edges $E$ is built connecting two vertices $v_i$ and $v_j$, which represent two pixels $i$ and $j$, by a directed edge from $v_i$ to $v_j$, $e_{v_i,v_j} \in E$, if the Euclidean distance between them is less than or equal to a radius $r$ and $I(i) \leqslant I(j)$, according to

\begin{equation}
E = \left \{e_{v_i,v_j} \in E \mid dist(v_i,v_j) \leqslant r   \wedge  I(i) < I(j) \right \},
\end{equation}
where $dist(v_i,v_j) = \sqrt{(x_i-x_j)^2 + (y_i-y_j)^2}$ is the Euclidean distance between two pixels. 
Each edge has a weight $w(e_{v_i,v_j})$ defined as   

\begin{equation} 
 w(e_{v_i,v_j}) = 
\left\{\begin{matrix}
 \frac{\mid I(i) - I(j) \mid}{L}, & \textnormal{If } r=1  \\ 
\frac{ \left( \frac{dist(v_i,v_j)-1}{r-1} \right) + \left(\frac{\mid I(i) - I(j) \mid}{L} \right)}{2}, & \textnormal{Otherwise.}
\end{matrix}\right.
\end{equation}

\begin{figure}[!ht]
	\centering
	\subfigure[Texture image]{\includegraphics[width=0.25\textwidth]{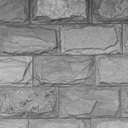}}\\
	\subfigure[$r=1$]{\includegraphics[width=0.30\textwidth]{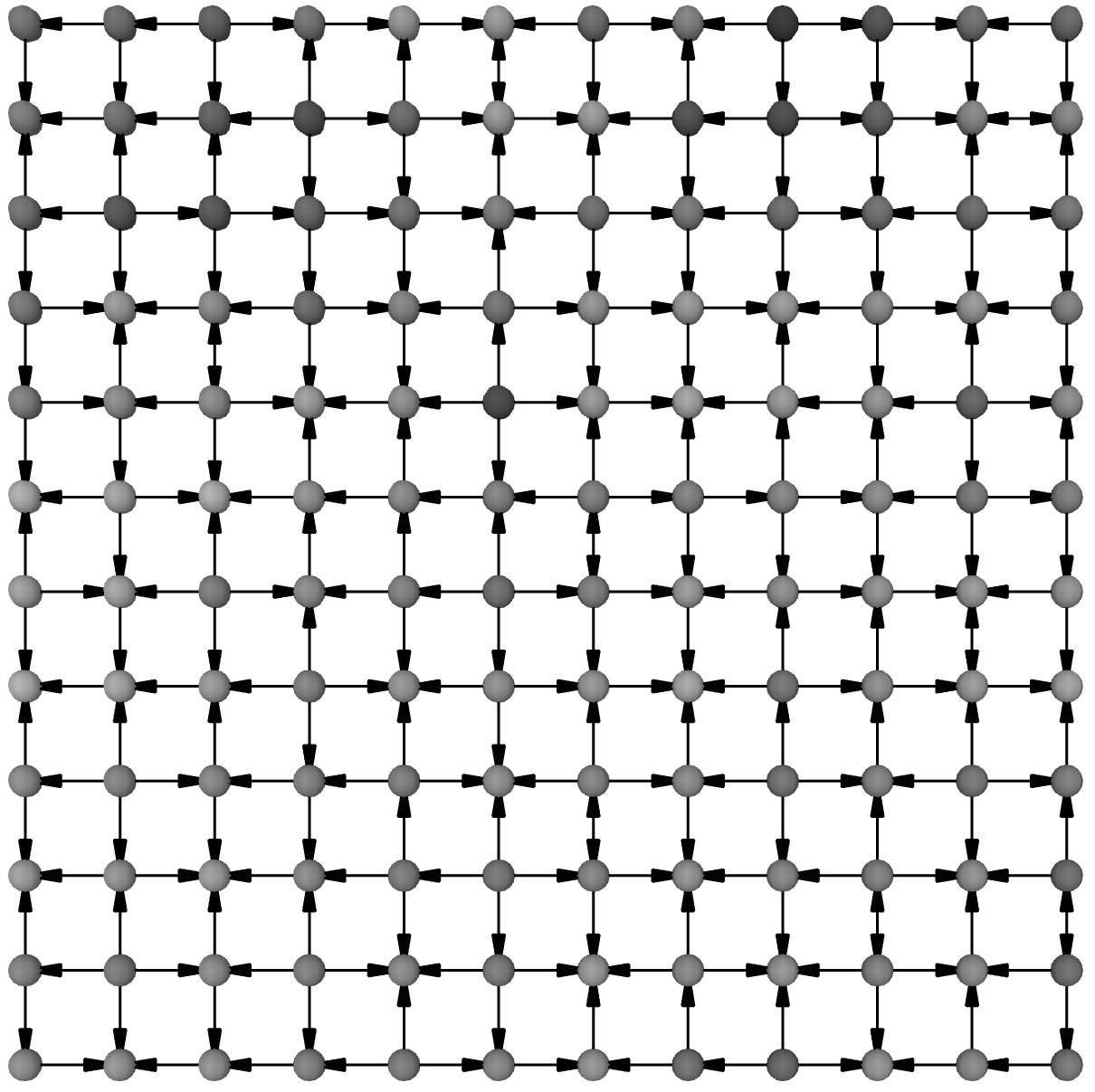}} \hspace{8mm}
    \subfigure[$r=2$]{\includegraphics[width=0.30\textwidth]{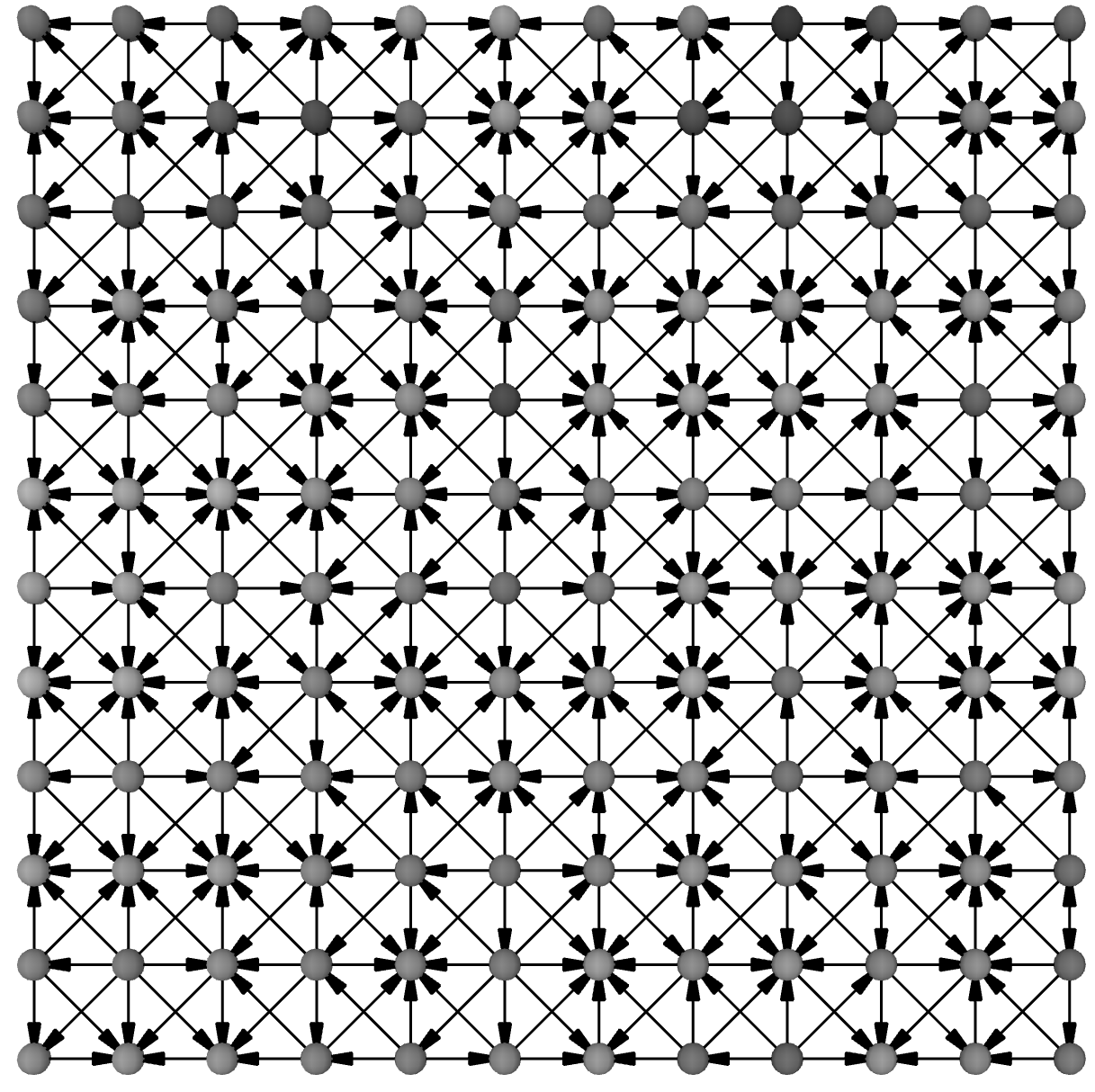}}\\
	\caption{Examples of a texture image modeled as a directed complex network.}
	\label{fig:model_dir_gray}
\end{figure}

It is worth mentioning that the direction of the edges is determined by the pixel gray-levels.
In other words, an edge points to the vertex that represents a pixel with greater intensity. If both intensities are equal, the edge is bidirectional. 
It is also important to stress that the value $r$ is the unique parameter of modeling and determines the size of the neighborhood of each vertex. Thus, as $r$ increases, the reach of connection and the degree of vertices increase as well. This procedure makes the analysis of the behavior of this evolution an interesting way of studying these networks. Figure \ref{fig:model_dir_gray} shows the modeling of a texture image as a directed network for different values of $r$.

\subsection{Proposed signature based on RNN} 

Our method aims to use as texture signature the weights of the output layer of the RNN trained with information from the modeled complex networks. 
For this, three sources of information are considered for each vertex: out-degree, weighted out-degree, and weighted in-degree. As the out-degree is directly related to the in-degree in the modeled networks (i.e. the sum of these two degrees is equal in all the vertices) and, therefore, provide the same information, we considered only the out-degree.      

The out-degree $k_{v_i}$ of a vertex $v_i$ represents the number of out-edges connected to other vertices, 

\begin{equation}
k_{v_i} = \sum_{v_j \in V}^{} \left\{\begin{array}{rcl}
	1,& e_{v_i,v_j} \in E \\	
    0,& \textnormal{otherwise}.
	\end{array}\right.
\end{equation}

On the other hand, the weighted out-degree $ks_{v_i}$ is given by the sum of the weights of the out-degree edges of a vertex $v_i$,

\begin{equation}
ks_{v_i} = \sum_{v_j \in V}^{} \left\{\begin{array}{rcl}
	w(e_{v_i,v_j}) ,& e_{v_i,v_j}\in E \\	
    0,& \textnormal{otherwise}.
	\end{array}\right.
\end{equation}

Finally, the weighted in-degree $ke_{v_i}$ is defined as the sum of the weights of the in-degree edges in $v_i$, 

\begin{equation}
ke_{v_i} = \sum_{v_j \in V}^{} \left\{\begin{array}{rcl}
	w(e_{v_j,v_i}) ,& e_{v_j,v_i}\in E \\	
    0,& \textnormal{otherwise}.
	\end{array}\right.
\end{equation}

\begin{figure}[!ht]
	\centering
	\includegraphics[width=0.8\textwidth]{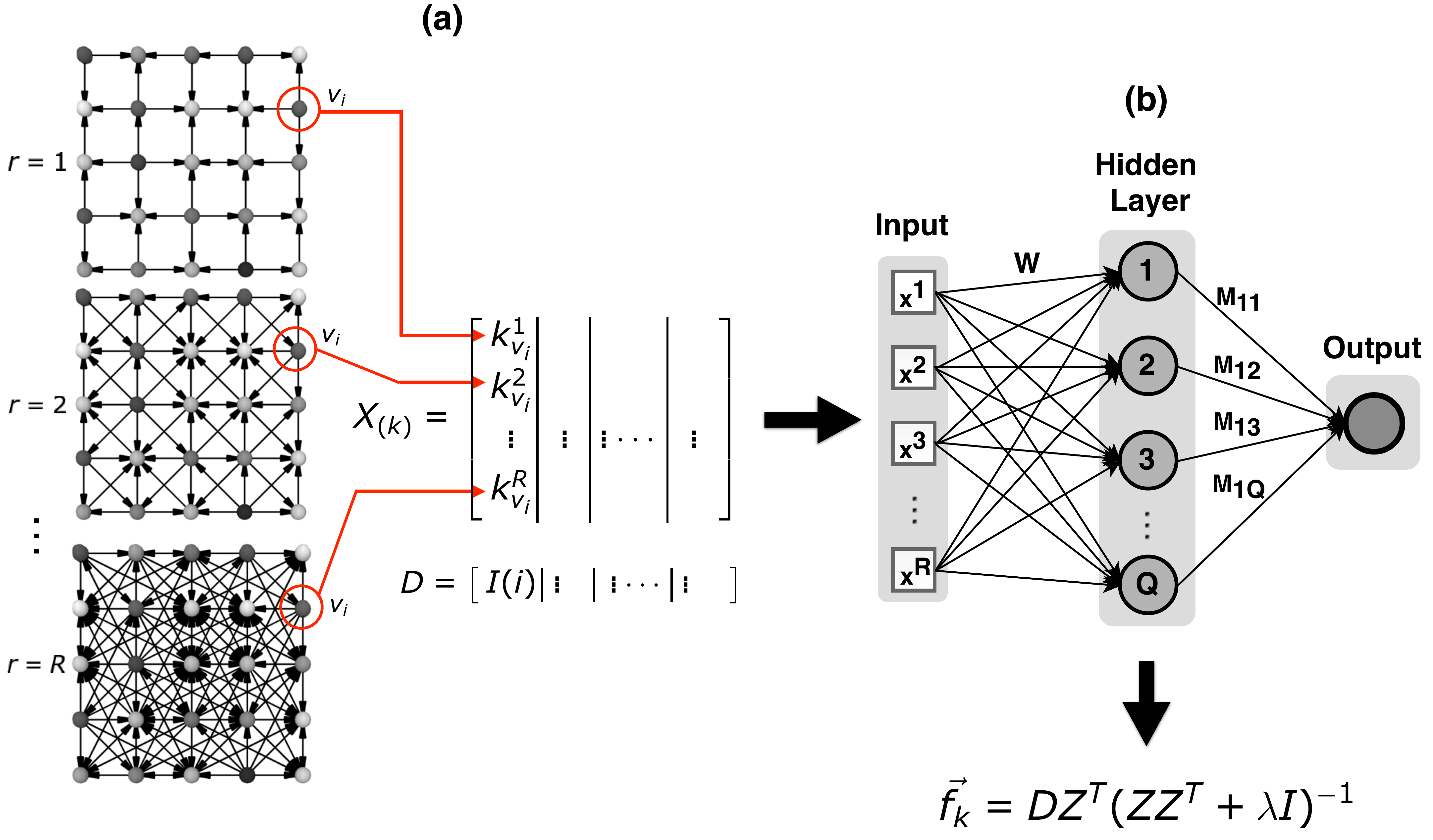}
    \caption{Building of an input feature vector and corresponding output label for the out-degree $k_{v_i}$ using different values of $r$ to model the complex networks.}
	\label{fig:XeD}
\end{figure}

To build a matrix of input vector for the RNN, we adopted a strategy of analysis of the evolution of the complex network for different values of the modeling parameter $r$. In this way, the input feature vector and the corresponding label of a vertex $v_i$ are built according to the following procedure: the gray-scale intensity of the pixel is considered as an output label $d_{v_i} = I(i)$ and the values of out-degree of the vertex for different values of the modeling parameter $r$ are attributes of the input feature vector $\vec{x}_{v_i} = [k_{v_i}^1, k_{v_i}^2, ..., k_{v_i}^R]$, where $R$ is the maximum values of the modeling parameter.
A matrix of input feature vectors $X_{(k)}$ and a matrix of output labels $D$ are then built for all the vertices of the complex network. Thus, it is possible to analyze the evolution of the topology of vertices that represent pixels that have a determined gray-scale intensity. Figure \ref{fig:XeD}(a) shows an example of how to build these matrices $X$ and $D$. In addition to building the matrix of input feature vectors $X_{(k)}$ for the out-degree, we also built matrices of input vectors for the weighted out-degree $X_{(ks)}$ and for the weighted in-degree $X_{(ke)}$.

The next step is to define the weights of the matrix $W$ of the hidden layer of the RNN. In general, these weights are determined randomly in each training stage. Nevertheless, because we want our method to provide the same signature for the same texture image, it is necessary to use the same values in the matrix $W$. Thus, we adopted the strategy proposed in \cite{JarbasRNN2015} and used the Linear Congruent Generator (LCG) \cite{rLEH51a,park1988random} in order to obtain pseudo-random uniform values for the matrix $W$, according to the following equation

\begin{equation}
V(n+1) = (a*V(n)+b) \textnormal{ mod } c,
\end{equation}
where $V$ is a random numeric sequence and $a$, $b$ and $c$ are parameters. The sequence $V$ has length $E = Q*(p + 1)$, its first value is $V(1) = E + 1$, and the values of the parameters are $a=E + 2$, $b = E + 3$ and $c = E^2$ (values adopted in \cite{JarbasRNN2015}). Hence, the matrix $W$ is composed of the vector $V$ divided into $Q$ segments of length $p+1$. Finally, all values of matrix $W$ and each line of the matrix $X$ are normalized using z-score (zero mean and unit variance).

The proposed texture signature is built based on the matrix $M$, which becomes a vector $\vec{f} = DZ^T(ZZ^T+\lambda I)^{-1}$, where $\lambda=10^{-3}$ (Figure \ref{fig:XeD}(b)). Notice that $\vec{f}$ has  length $Q + 1$ due to the bias weight. Thus, the first step is to concatenate the vectors $\vec{f}$ obtained from RNNs trained with the three matrices of input data $X_{(k)}$, $X_{(ks)}$, $X_{(ke)}$, according to

\begin{equation}
\vec{\Upsilon}(Q)_R = \left[ \vec{f_{k}}, \vec{f_{ks}}, \vec{f_{ke}} \right],
\end{equation}
where $Q$ is the number of neurons of the hidden layer and $R$ is the maximum radius for building the complex network.

The vector $\vec{\Upsilon}(Q)_R$ is built using a single value of $Q$ and $R$. These two parameters influence the weights of the neural network and, therefore, provide different characteristics for different values. Thus, initially we propose a vector $\vec{\Theta}(R)_{(Q_1, Q_2, Q_m)}$ that concatenates the vectors $\vec{\Upsilon}(Q)_R$ for different values of $Q$,

\begin{equation}
\vec{\Theta}(R)_{Q_1, Q_2, ..., Q_m} = \left[ \vec{\Upsilon}(Q_1)_R, \vec{\Upsilon}(Q_2)_R, ..., \vec{\Upsilon}(Q_m)_R \right].
\end{equation}

Finally, we propose a feature vector $\vec{\Psi}(R_1, R_2)_{Q_1, Q_2, ..., Q_m}$ that concatenates the vector $\vec{\Theta}(R)_{Q_1, Q_2, ..., Q_m}$ for two values of $R$,

\begin{equation}
\vec{\Psi}(R_1, R_2)_{Q_1, Q_2, ..., Q_m} = \left[ \vec{\Theta}(R_1)_{Q_1, Q_2, ..., Q_m}, \vec{\Theta}(R_2)_{Q_1, Q_2, ..., Q_m} \right].
\end{equation}

\section{Experiments} \label{sec:experiments}

In order to validate our proposed method and compare it to other texture analysis methods, the signatures were classified using linear discriminant analysis \cite{fukunaga-1990}. This classifier was adopted due to its simplicity, which emphasizes the characteristics obtained by the methods. The leave-one-out cross-validation scheme was used. In this validation strategy, one sample is used for testing the model and the remainder for training it. This process is repeated $N$ times ($N$ is the number of samples), each time with a different sample for testing. The performance measure is the average accuracy of the $N$ runnings.

The gray-scale texture databases used as benchmark to evaluate our proposed method were:

\begin{itemize}

\item Brodatz \cite{brodatz-1966}: just as in \cite{cndt}, 1776 texture images of 128 $\times$ 128 pixel size from this database divided into 111 classes were used in this work.  

\item Outex \cite{OjalaMPVKH02}: just as in \cite{cndt}, the original 68 images $746 \times 538$ from TC\_Outex\_00013 were divided into 20 sub-images 128 $\times$ 128 pixel size without overlapping. Thus, the database used in this work has 1360 textures.  

\item USPTex \cite{backes2012}: this database has 2292 samples divided into 191 classes, 12 images per class, and each image has 128 $\times$ 128 pixel size.

\item Vistex: the database \textit{Vision Texture} is provided by the Vision and Modeling Group - MIT Media Lab \cite{Vistex1995}. Just as in \cite{cndt}, the original 54 images $512 \times 512$ were split into 16 sub-images 128 $\times$ 128 pixel size without overlapping. Thus, the database used in this work has 864 images.

\end{itemize}

The proposed method is applied to the aforementioned databases and the accuracy is compared to other methods of the literature. They are: Grey-Level Co-occurrence Matrix (GLCM) \cite{haralick1973,haralick1979statistical}, Gray Level Difference Matrix (GLDM) \cite{weszka1976comparative,kim1999statistical}, Fourier \cite{azencott1997texture}, Gabor Filters \cite{manjunath1996texture,idrissa2002texture}, Fractal \cite{backes2009plant}, Fractal Fourier \cite{florindo2012fractal}, Local Binary Patterns (LBP) \cite{ojala2002multiresolution}, Local Binary Patterns Variance (LBPV) \cite{lbpv}, Complete Local Binary Pattern (CLBP) \cite{clbp}, Local Phase Quantization (LPQ) \cite{lpq}, Local Configuration Pattern (LCP) \cite{lcp}, Local Frequency Descriptor (LFD) \cite{lfd}, Binarized Statistical Image Features (BSIF) \cite{kannala2012bsif}, Local Oriented Statistics Information Booster (LOSIB) \cite{losib}, Adaptive Hybrid Pattern (AHP) \cite{zhu2015adaptive}, Complex Network Texture Descriptors (CNTD) \cite{cndt} and ELM signature \cite{JarbasRNN2015}.   

\section{Results and Discussion} \label{sec:results}
\subsection{Parameter Evaluation}

\begin{figure}[!ht]
 \label{fig:valuesQ}
 \centering
 \includegraphics[width=0.49\textwidth]{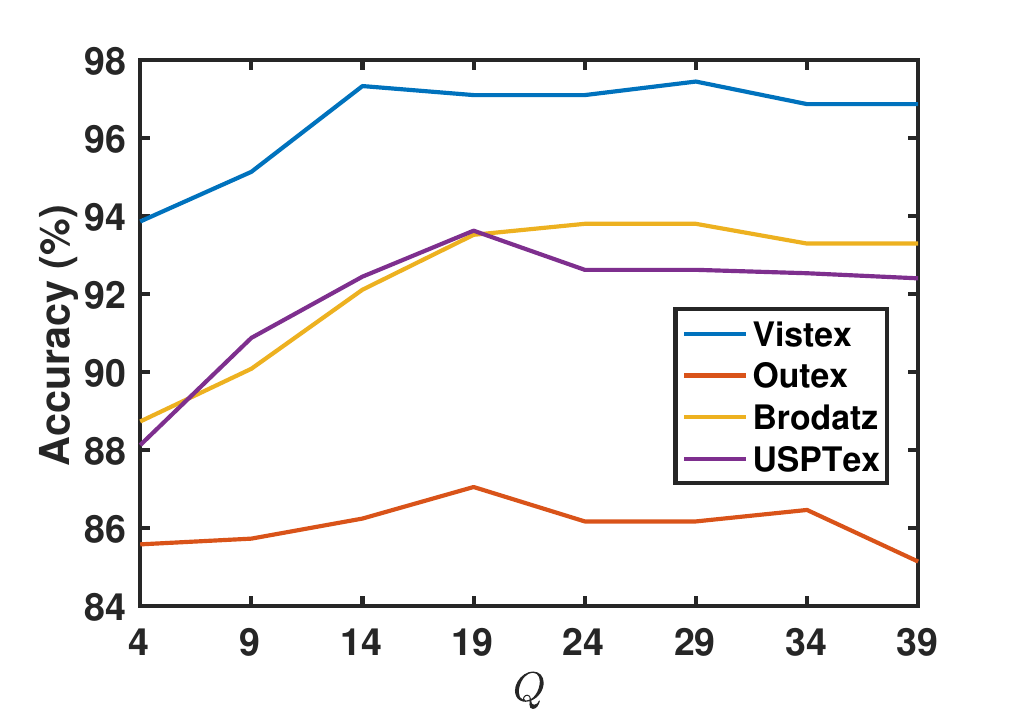}
 \caption{Accuracies of the feature vector $\vec{\Upsilon}(Q)_R$ using different values of $Q$ on the four databases.}
\end{figure}

Figure \ref{fig:valuesQ} shows the accuracies achieved on the four databases with the feature vector $\vec{\Upsilon}(Q)_R$ using $R=4$.
In this experiment, we used different values of $Q \in \{04, 09, 14, 19, 29, 39\}$, which were selected because they produce a number of features that is multiple of five for each feature vector considered.
As can be seen in the figure, the success rates increase as we increase the value of $Q$.
This increase is followed by an increase in the number of features used.
The best accuracies are obtained using $Q=14$ on the Vistex database and $Q = 19$ on the other databases.
These values of $Q$ produce feature vectors of size 45 ($Q=14$) and 60 ($Q=19$).
Furthermore, the success rate stabilizes when we use values of $Q$ larger than $Q = 14$ on the Vistex database and values larger than $Q=19$ on the other databases.

Table \ref{tab:evaluate_q_texture} shows the accuracies obtained on the four databases using the feature vector $\vec{\Theta}(R)_{Q_1, Q_2, ..., Q_m}$ with $R=4$. 
The results show that as the values of $Q$ and its combinations increase (i.e. the number of features increases), the success rates increase as well. However, very large feature vectors do not assure the highest performance, once the success rates tend to stabilize at a determined value. For instance, if we compare the vector $\vec{\Theta}(04)_{19, 29, 39}$, which has 270 features, with the vector $\vec{\Theta}(04)_{04, 09, 14}$, which has 90 attributes, the former has a lower performance in all the databases. This suggests that the proposed signature reaches its limit in terms of discrimination. Thus, we considered the vectors $\vec{\Theta}(04)_{04, 09, 14}$ and $\vec{\Theta}(04)_{04, 14, 19}$, since they presented a good trade-off between high accuracy and a small number of features.

\begin{table}[!ht]                                              
\centering         
\caption{Accuracies of the feature vector $\vec{\Theta}(R)_{Q_1, Q_2, ..., Q_m}$ using different values of $Q$ and their combinations for the maximum radius $R=4$.}

\begin{tabular}{cccccc} \hline                                    
\textbf{$\{Q_1, Q_2, ..., Q_m\}$} & \textbf{No of features} & \textbf{Outex} & \textbf{USPTex} & \textbf{Brodatz} & \textbf{Vistex} \\ \hline
      
\{04, 09\} & 45 & 88.60 & 94.07 & 93.02 & 97.22 \\           
\{04, 14\} & 60 & 88.82 & 94.98 & 93.86 & 98.50 \\          
\{04, 19\} & 75 & 88.97 & 95.46 & 94.88 & 97.92 \\          
\{04, 29\} & 105 & 88.38 & 94.76 & 94.48 & 97.80 \\         
\{04, 39\} & 135 & 87.57 & 95.42 & 95.27 & 98.15 \\         
\{09, 14\} & 75 & 88.09 & 94.72 & 93.52 & 97.92 \\          
\{09, 19\} & 90 & 88.60 & 94.42 & 94.26 & 98.15 \\          
\{09, 29\} & 120 & 87.50 & 94.24 & 94.14 & 97.80 \\         
\{09, 39\} & 150 & 86.76 & 94.81 & 94.88 & 97.80 \\         
\{14, 19\} & 105 & 89.04 & 94.94 & 94.54 & 98.26 \\        
\{14, 29\} & 135 & 88.09 & 94.90 & 94.48 & 97.92 \\        
\{14, 39\} & 165 & 87.65 & 95.29 & 95.33 & 98.15 \\        
\{19, 29\} & 150 & 87.50 & 94.59 & 94.76 & 97.45 \\        
\{19, 39\} & 180 & 87.28 & 95.03 & 95.27 & 98.03 \\        
\{29, 39\} & 210 & 85.88 & 94.37 & 95.05 & 97.92 \\        
\textbf{\{04, 09, 14\}} & \textbf{90} & \textbf{89.34} & \textbf{95.46} & \textbf{95.05} & \textbf{98.61} \\      
\{04, 09, 19\} & 105 & 89.71 & 95.55 & 95.16 & 98.26 \\     
\{04, 09, 29\} & 135 & 88.68 & 95.51 & 94.88 & 97.80 \\     
\{04, 09, 39\} & 165 & 87.94 & 95.90 & 95.72 & 97.80 \\     
\textbf{\{04, 14, 19\}} & \textbf{120} & \textbf{89.41} & \textbf{95.94} & \textbf{95.21} & \textbf{98.38} \\     
\{04, 14, 29\} & 150 & 88.68 & 95.59 & 95.05 & 98.26 \\     
\{04, 14, 39\} & 180 & 88.53 & 95.90 & 95.89 & 98.84 \\     
\{04, 19, 29\} & 165 & 89.12 & 95.94 & 95.27 & 98.03 \\     
\{04, 19, 39\} & 195 & 88.68 & 95.90 & 95.61 & 98.38 \\     
\{04, 29, 39\} & 225 & 87.94 & 95.03 & 95.72 & 98.38 \\     
\{09, 14, 19\} & 135 & 89.56 & 95.24 & 94.99 & 98.50 \\     
\{09, 14, 29\} & 165 & 88.75 & 95.42 & 95.10 & 98.15 \\     
\{09, 14, 39\} & 195 & 88.01 & 95.68 & 95.61 & 98.73 \\     
\{09, 19, 29\} & 180 & 88.75 & 95.24 & 94.93 & 98.26 \\     
\{09, 19, 39\} & 210 & 87.94 & 95.38 & 95.05 & 97.92 \\     
\{09, 29, 39\} & 240 & 88.01 & 94.59 & 94.99 & 97.92 \\     
\{14, 19, 29\} & 195 & 88.75 & 95.33 & 95.05 & 98.50 \\    
\{14, 19, 39\} & 225 & 88.38 & 95.72 & 95.44 & 98.61 \\    
\{14, 29, 39\} & 255 & 88.09 & 95.03 & 95.50 & 98.50 \\    
\{19, 29, 39\} & 270 & 88.01 & 94.63 & 95.50 & 97.92 \\ 
\hline
\end{tabular} 

\label{tab:evaluate_q_texture}                                 
\end{table}

\begin{figure}[!ht]
	\centering
	\subfigure[$Q = \{04, 09, 14\}$]{\includegraphics[width=0.49\textwidth]{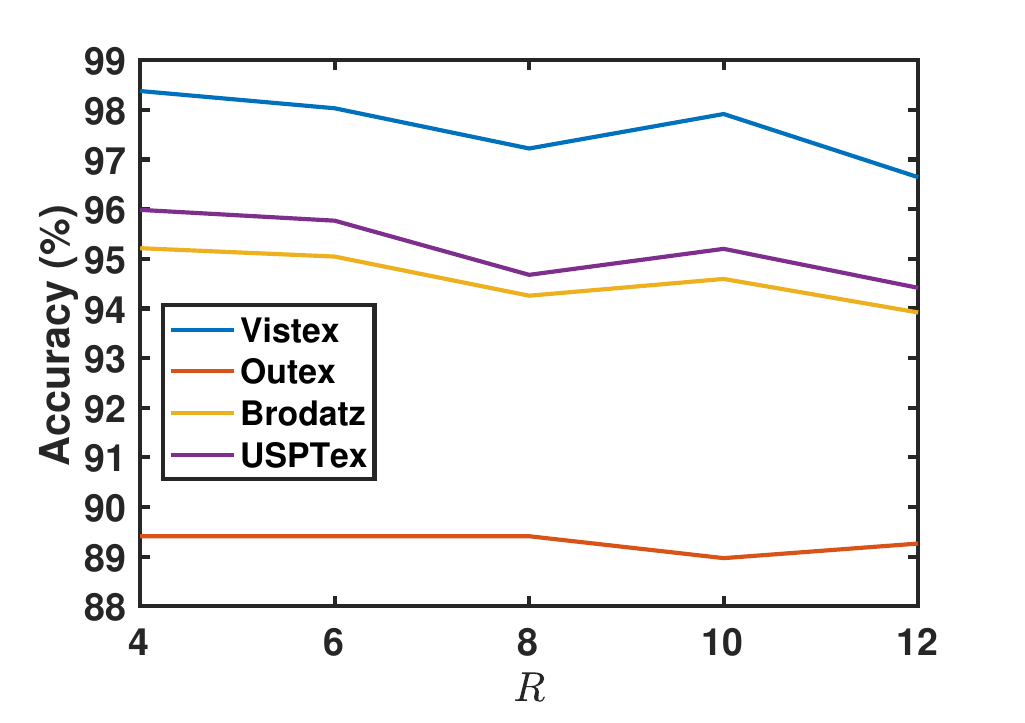}}
	\subfigure[$Q = \{04, 14, 19\}$]{\includegraphics[width=0.49\textwidth]{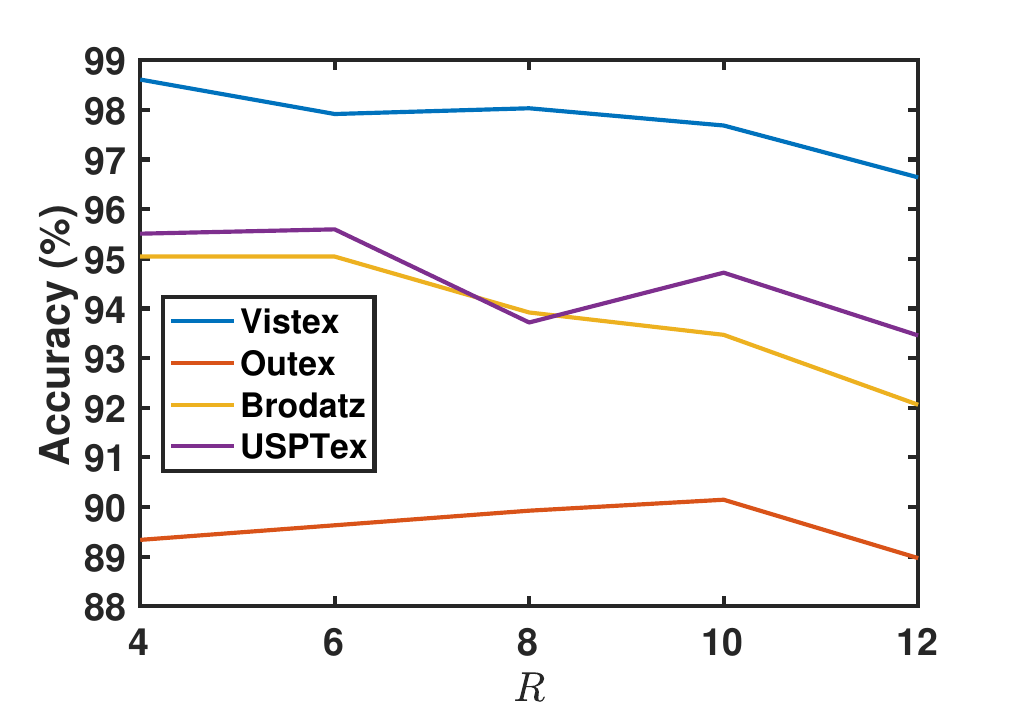}}\\
	\caption{Accuracies using the feature vector $\vec{\Theta}(R)_{Q_1, Q_2, ..., Q_m}$ for the two better set of $Q$ with different values of maximum radius $R$.}
	\label{fig:values_R}
\end{figure}

We also evaluated the feature vector $\vec{\Theta}(R)_{Q_1, Q_2, ..., Q_m}$ for different values of $R$.
Figure \ref{fig:values_R} shows the accuracies yielded considering the combinations $Q=\{04, 09, 14\}$ and $Q=\{04, 14, 19\}$. 
The value of maximum radius $R$ is associated to the zone of connection between the pixels (i.e., vertices). Thus, lower values of $R$ represent the closest pixels and, as $R$ increases, the reach of connection increases as well.
The results show that the lowest values of $R$ provide better accuracies when compared to the highest values.
This demonstrates that local patterns are more important than global patterns to discriminate the textures.

Furthermore, we analyzed the combination of vectors $\vec{\Theta}(R)_{Q_1, Q_2, ..., Q_m}$ (showed in Table \ref{tab:evaluate_q_texture}) with different values of maximum radius $R$, resulting in the vector $\vec{\Psi}(R_1, R_2)_{Q_1, Q_2, ..., Q_m}$. To compute this vector, we used the combinations of $Q$ that provided the best results in Table \ref{tab:evaluate_q_texture}: $Q=\{4, 9, 14\}$ and $Q = \{4, 14, 19\}$. In this experiment, we computed the vector $\vec{\Psi}(R_1, R_2)_{Q_1, Q_2, ..., Q_m}$ for two values of $R$ (i.e, up to two combinations of $\vec{\Theta}(R)_{Q_1, Q_2, ..., Q_m}$) due to the large number of features generated.

Table \ref{tab:vetR1} shows the results of the vectors $\vec{\Psi}(R_1, R_2)_{Q_1, Q_2, ..., Q_m}$ using the combination $Q = \{04, 09, 14\}$. 
The highest accuracy was provided by the vector $\vec{\Psi}(04, 06)_{04, 09, 14}$. 
The results of the vectors $\vec{\Psi}(R_1, R_2)_{Q_1, Q_2, ..., Q_m}$ built with the combination $Q = \{04, 14, 19\}$ are shown in Table \ref{tab:vetR2}. 
In this experiment, the best results were obtained by using the vector $\vec{\Psi}(04, 10)_{04, 14, 19}$. 
Tables \ref{tab:vetR1} and \ref{tab:vetR2} also show that by combining the vector $\vec{\Theta}(R)_{Q_1, Q_2, ..., Q_m}$ with different values of $R$, the accuracy increases approximately 1\% on the databases.
However, in the two cases, the combinations of high values of $R$ provide inferior results.
Even though the best results of the two Tables ($\vec{\Psi}(04, 06)_{04, 09, 14}$ and $\vec{\Psi}(04, 10)_{04, 14, 19}$) are similar, notice that the vector $\vec{\Psi}(04, 10)_{04, 14, 19}$ has a number of features larger than the vector $\vec{\Psi}(04, 06)_{04, 09, 14}$.

\begin{table}[!ht]                                              
\centering          
\caption{Accuracies using different sets of radii $R$ and $Q = \{04, 09, 14\}$.}
\begin{tabular}{cccccc} \hline                                  
$\{R_1, R_2\}$ & No of features & Outex & USPTex & Brodatz & Vistex \\\hline
     
\textbf{\{04, 06\}} & \textbf{180} & \textbf{91.54} & \textbf{96.64} & \textbf{96.11} &\textbf{ 98.73} \\          
\{04, 08\} & 180 & 91.47 & 96.25 & 95,88 & 98.26 \\          
\{04, 10\} & 180 & 91.47 & 96.55 & 95.83 & 98.84 \\         
\{04, 12\} & 180 & 91.69 & 96.25 & 95.72 & 98.26 \\         
\{06, 08\} & 180 & 91.54 & 96.47 & 95.77 & 98.61 \\          
\{06, 10\} & 180 & 90.74 & 96.25 & 95.83 & 98.50 \\         
\{06, 12\} & 180 & 90.44 & 96.38 & 95.83 & 98.38 \\         
\{08, 10\} & 180 & 90.58 & 96.03 & 95.15 & 98.49 \\         
\{08, 12\} & 180 & 90.29 & 95.72 & 95.21 & 98.14 \\         
\{10, 12\} & 180 & 90.59 & 95.46 & 94.93 & 98.38 \\ \hline        
\end{tabular}                                              
\label{tab:vetR1}                                 
\end{table}     

\begin{table}[!ht]                                              
\centering    
\caption{Accuracies using different sets of radii $R$ and $Q = \{4, 14, 19\}$.}

\begin{tabular}{cccccc}   \hline                                 
$\{R_1, R_2\}$ & No of features & Outex & USPTex & Brodatz & Vistex \\ \hline
          
\{04, 06\} & 240 & 90.07 & 96.73 & 95.83 & 99.19 \\          
\{04, 08\} & 240 & 91.17 & 96.68 & 96.05 & 98.95 \\          
\textbf{\{04, 10\}} & \textbf{240} & \textbf{91.32} & \textbf{96.95} & \textbf{96.06} &\textbf{ 99.19} \\         
\{04, 12\} & 240 & 91.76 & 96.64 & 96.11 & 98.61 \\         
\{06, 08\} & 240 & 90.14 & 96.29 & 96.39 & 98.61 \\          
\{06, 10\} & 240 & 89.63 & 96.60 & 95.95 & 98.50 \\         
\{06, 12\} & 240 & 90.29 & 96.55 & 96.06 & 98.26 \\         
\{08, 10\} & 240 & 90.00 & 95.94 & 95.72 & 98.37\\         
\{08, 12\} & 240 & 91.32 & 95.94 & 95.77 & 98.03 \\         
\{10, 12\} & 240 & 90.51 & 96.03 & 95.05 & 98.03 \\ \hline      
\end{tabular}                                              
\label{tab:vetR2}                                  
\end{table}

\begin{table}[!ht]  
\centering  
\caption{Comparison of accuracies of different texture analysis methods in four texture databases. A subset of the compared results is present in \cite{JarbasRNN2015}}.  
\begin{tabular}{lrcccc} \hline                               
Methods & No of features & Outex & USPTex & Brodatz & Vistex \\ \hline
GLCM \cite{haralick1979statistical}  &  24 & 80.73 & 83.64 & 90.43 & 92.24 \\ 
GLDM \cite{weszka1976comparative}  & 60 & 86.76 & 92.06 & 94.43 & 97.11 \\  
Gabor Filters \cite{manjunath1996texture}  & 48 & 81.91 &  89.22 & 89.86 & 93.29 \\ 
Fourier \cite{weszka1976comparative}  & 63 & 81.91 & 67.50 & 75.90 & 79.51 \\    
Fractal \cite{backes2009plant}   & 69 & 80.51 & 78.27 & 87.16 & 91.67 \\ 
Fractal Fourier \cite{florindo2012fractal}   & 68 & 68.38 & 59.47 & 71.96 & 79.75 \\ 
LBP \cite{ojala2002multiresolution} & 256 & 81.10 & 85.43 & 93.64 & 97.92 \\                      
LBPV \cite{lbpv}  & 555 & 75.66 & 54.97 & 86.26 & 88.65 \\
CLBP \cite{clbp} & 648 & 85.80 & 91.14 & 95.32 & 98.03 \\                       
AHP \cite{zhu2015adaptive} & 120 & 88.31 & 94.85 & 94.88 & 98.38 \\    
BSIF \cite{kannala2012bsif}  & 256 & 77.43 & 77.66 & 91.44 & 88.66 \\                     
LCP \cite{lcp}  & 81 & 86.25 & 91.14 & 93.47 & 94.44 \\                       
LFD \cite{lfd} & 276 & 82.57 & 83.55 & 90.99 & 94.68 \\                      
LPQ \cite{lpq} & 256 & 79.41 & 85.12 & 92.51 & 92.48 \\                      
ELM Signature \cite{JarbasRNN2015}  & 180 & 89.71 & 95.11 & 95.27 & 98.15 \\ 
CNTD \cite{cndt}  & 108  & 86.76 & 91.71 & 95.27 & 98.03 \\
\hline  
$\vec{\Theta}(04)_{04, 09, 14}$ & 90 & 89.34 & 95.46 & 95.05 & 98.61 \\      
$\vec{\Psi}(04, 06)_{04, 09, 14}$ & 180 & \textbf{91.54} & 96.64 & \textbf{96.11} & 98.73 \\  
$\vec{\Psi}(04, 10)_{04, 14, 19}$ & 240 & 91.32 & \textbf{96.95} & 96.06 &\textbf{ 99.19} \\ \hline        
\end{tabular}                                                     
\label{tab:comp_textures}                                        
\end{table} 

\subsection{Comparison with other methods}

To evaluate the results obtained by our proposed method, we performed comparisons with methods present in the literature. The experimental setup used was the same for all the methods (LDA with leave-one-out), except for CLBP, which used the classifier 1-Nearest Neighborhood (1-NN) with distance chi-square, according to the original paper. For our method, we adopted the two texture signatures that obtained the best results in the previous analysis: $\vec{\Psi}(04, 06)_{04, 09, 14}$ and $\vec{\Psi}(04, 10)_{04, 14, 19}$.

Table \ref{tab:comp_textures} presents the results obtained by all the methods in the four image databases evaluated. The results show that our proposed method obtained the best results when compared to the other methods using both signatures. Also, it is important to stress that our method reached higher accuracies than the ELM signature and CNDT method (which is also based on complex networks). This suggests that our method obtained superior performance because it has simultaneously the main characteristics of both compared methods. In other words, the ELM signature uses only pixel intensities to train the neural network, without any valuable information from complex network modeling, and the CNTD method models images as complex networks and computes only traditional measures, without using a neural network to extract the deep characteristics from these complex networks.

Even though our proposed method has signatures with a larger number of descriptors when compared to some methods of the literature, it is important to emphasize that, if we consider only the vector $\vec{\Theta}(R)_{Q_1, Q_2, ..., Q_m}$, the results are still competitive. For instance, the vector  $\vec{\Theta}(04)_{4, 9, 14}$, which has only 90 features, provides superior performance on the Vistex and USPTex databases. In the remainder databases, the results are very close to the highest accuracies (only 0.37\% smaller than the result of ELM signature on the Outex database and 0.27\% smaller than the accuracy of the CLBP on the Brodatz database)

\section{Conclusion} \label{sec:conclusion}

This paper presented an innovative approach of texture feature extraction based on the fusion of complex network and randomized neural network.
In the proposed method, a new approach to model the image as a CN that uses only a parameter is presented.
We also proposed a new way of characterizing the CN based on the idea of using the output weights of a randomized neural network trained with topological properties of the CN.
The obtained classification results on four databases outperformed other texture literature methods.
Also, the proposed approach has an excellent trade-off between performance and size of the feature vectors.
This demonstrates that the proposed approach is highly discriminative using the three feature vectors considered.
In this way, this paper shows that the fusion of complex network and randomized neural network is a research field with great potential as a feasible texture analysis methodology.

\section*{Acknowledgments}

 Lucas Correia Ribas gratefully acknowledges the financial support grant \#2016/23763-8, S\~ao Paulo Research Foundation (FAPESP).
Jarbas Joaci de Mesquita S\'a Junior thanks CNPq (National Council for Scientific and Technological Development, Brazil) (Grant: 302183/2017-5) for the financial support of this work. Leonardo Felipe dos Santos Scabini acknowledges support from CNPq (Grant \#134558/2016-2).
Odemir M. Bruno thanks the financial support of CNPq (Grant \# 307797/2014-7) and FAPESP (Grant \#s 14/08026-1 and 16/18809-9).

\bibliographystyle{acm}  
\bibliography{references}  

\begin{thebibliography}{10}

\bibitem{azencott1997texture}
{\sc Azencott, R., Wang, J.-P., and Younes, L.}
\newblock Texture classification using windowed {F}ourier filters.
\newblock {\em IEEE Transactions on Pattern Analysis and Machine Intelligence
  19}, 2 (1997), 148--153.

\bibitem{backes2008fractal}
{\sc Backes, A.~R., and Bruno, O.~M.}
\newblock A new approach to estimate fractal dimension of texture images.
\newblock In {\em International Conference on Image and Signal Processing\/}
  (2008), Springer, pp.~136--143.

\bibitem{backes2009plant}
{\sc Backes, A.~R., Casanova, D., and Bruno, O.~M.}
\newblock Plant leaf identification based on volumetric fractal dimension.
\newblock {\em International Journal of Pattern Recognition and Artificial
  Intelligence 23}, 06 (2009), 1145--1160.

\bibitem{backes2012}
{\sc Backes, A.~R., Casanova, D., and Bruno, O.~M.}
\newblock Color texture analysis based on fractal descriptors.
\newblock {\em Pattern Recognition 45}, 5 (2012), 1984--1992.

\bibitem{cndt}
{\sc Backes, A.~R., Casanova, D., and Bruno, O.~M.}
\newblock Texture analysis and classification: A complex network-based
  approach.
\newblock {\em Information Sciences 219\/} (2013), 168--180.

\bibitem{barabasi2016network}
{\sc Barab{\'a}si, A.~L.}
\newblock {\em Network science}.
\newblock Cambridge university press, 2016.

\bibitem{scalefreeCN}
{\sc Barab{\'a}si, A.-L., and Albert, R.}
\newblock Emergence of scaling in random networks.
\newblock {\em science 286}, 5439 (1999), 509--512.

\bibitem{BHARATI200457}
{\sc Bharati, M.~H., Liu, J., and MacGregor, J.~F.}
\newblock Image texture analysis: methods and comparisons.
\newblock {\em Chemometrics and Intelligent Laboratory Systems 72}, 1 (2004),
  57 -- 71.

\bibitem{structureanddynamics}
{\sc Boccaletti, S., Latora, V., Moreno, Y., Chavez, M., and Hwang, D.-U.}
\newblock Complex networks: Structure and dynamics.
\newblock {\em Physics reports 424}, 4 (2006), 175--308.

\bibitem{brahnam2016local}
{\sc Brahnam, S., Jain, L.~C., Nanni, L., Lumini, A., et~al.}
\newblock {\em Local binary patterns: new variants and applications}.
\newblock Springer, 2016.

\bibitem{brodatz-1966}
{\sc Brodatz, P.}
\newblock {\em Textures: {A} photographic album for artists and designers}.
\newblock Dover Publications, New York, 1966.

\bibitem{BrunoPFC08}
{\sc Bruno, O.~M., de~Oliveira~Plotze, R., Falvo, M., and de~Castro, M.}
\newblock Fractal dimension applied to plant identification.
\newblock {\em INFORMATION SCIENCES 178}, 12 (2008), 2722--2733.

\bibitem{calvetti2000}
{\sc Calvetti, D., Morigi, S., Reichel, L., and Sgallari, F.}
\newblock Tikhonov regularization and the {L}-curve for large discrete
  ill-posed problems.
\newblock {\em Journal of Computational and Applied Mathematics 123}, 1 (2000),
  423 -- 446.

\bibitem{2013tumor}
{\sc Chicklore, S., Goh, V., Siddique, M., Roy, A., Marsden, P.~K., and Cook,
  G.~J.}
\newblock Quantifying tumour heterogeneity in 18f-fdg pet/ct imaging by texture
  analysis.
\newblock {\em European journal of nuclear medicine and molecular imaging 40},
  1 (2013), 133--140.

\bibitem{costa2007CNsurvey}
{\sc Costa, L. d.~F., Rodrigues, F.~A., Travieso, G., and Villas~Boas, P.~R.}
\newblock Characterization of complex networks: A survey of measurements.
\newblock {\em Advances in Physics 56}, 1 (2007), 167--242.

\bibitem{Cover1965}
{\sc Cover, T.~M.}
\newblock Geometrical and statistical properties of systems of linear
  inequalities with applications in pattern recognition.
\newblock {\em IEEE Transactions on Electronic Computers EC-14}, 3 (1965),
  326--334.

\bibitem{2004densesift}
{\sc Csurka, G., Dance, C., Fan, L., Willamowski, J., and Bray, C.}
\newblock Visual categorization with bags of keypoints.
\newblock In {\em ECCV International Workshop on Statistical Learning in
  Computer Vision\/} (2004), pp.~1--22.

\bibitem{cnAppSurvey2011}
{\sc da~Fontoura~Costa, L., Jr., O. N.~O., Travieso, G., Rodrigues, F.~A.,
  Boas, P. R.~V., Antiqueira, L., Viana, M.~P., and Rocha, L. E.~C.}
\newblock Analyzing and modeling real-world phenomena with complex networks: a
  survey of applications.
\newblock {\em Advances in Physics 60}, 3 (2011), 329--412.

\bibitem{CA2015texture}
{\sc da~Silva, N.~R., Van~der Wee{\"e}n, P., De~Baets, B., and Bruno, O.~M.}
\newblock Improved texture image classification through the use of a
  corrosion-inspired cellular automaton.
\newblock {\em Neurocomputing 149\/} (2015), 1560--1572.

\bibitem{JarbasRNN2015}
{\sc de~Mesquita~{Sá Junior}, J.~J., and Backes, A.~R.}
\newblock Elm based signature for texture classification.
\newblock {\em Pattern Recognition 51\/} (2016), 395 -- 401.

\bibitem{DEVES20142925}
{\sc de~Ves, E., Acevedo, D., Ruedin, A., and Benavent, X.}
\newblock A statistical model for magnitudes and angles of wavelet frame
  coefficients and its application to texture retrieval.
\newblock {\em Pattern Recognition 47}, 9 (2014), 2925 -- 2939.

\bibitem{florindo2012fractal}
{\sc Florindo, J.~B., and Bruno, O.~M.}
\newblock Fractal descriptors based on {F}ourier spectrum applied to texture
  analysis.
\newblock {\em Physica A: statistical Mechanics and its Applications 391}, 20
  (2012), 4909--4922.

\bibitem{dalcimar}
{\sc Florindo, J.~B., Casanova, D., and Bruno, O.~M.}
\newblock Fractal measures of complex networks applied to texture analysis.
\newblock In {\em Journal of Physics: Conference Series\/} (2013), vol.~410,
  IOP Publishing, p.~012091.

\bibitem{fukunaga-1990}
{\sc Fukunaga, K.}
\newblock {\em Introduction to Statistical Pattern Recognition}, 2nd~ed.
\newblock Academic Press, 1990.

\bibitem{losib}
{\sc Garc{\'\i}a-Olalla, O., Alegre, E., Fern{\'a}ndez-Robles, L., and
  Gonz{\'a}lez-Castro, V.}
\newblock Local oriented statistics information booster (losib) for texture
  classification.
\newblock In {\em Pattern Recognition (ICPR), 2014 22nd International
  Conference on\/} (2014), IEEE, pp.~1114--1119.

\bibitem{gonccalves2016texture}
{\sc Gon{\c{c}}alves, W.~N., da~Silva, N.~R., da~Fontoura~Costa, L., and Bruno,
  O.~M.}
\newblock Texture recognition based on diffusion in networks.
\newblock {\em Information Sciences 364\/} (2016), 51--71.

\bibitem{lcp}
{\sc Guo, Y., Zhao, G., and Pietik{\"a}inen, M.}
\newblock Texture classification using a linear configuration model based
  descriptor.
\newblock In {\em BMVC\/} (2011), Citeseer, pp.~1--10.

\bibitem{clbp}
{\sc Guo, Z., Zhang, L., and Zhang, D.}
\newblock A completed modeling of local binary pattern operator for texture
  classification.
\newblock {\em IEEE Transactions on Image Processing 19}, 6 (2010), 1657--1663.

\bibitem{lbpv}
{\sc Guo, Z., Zhang, L., and Zhang, D.}
\newblock Rotation invariant texture classification using lbp variance (lbpv)
  with global matching.
\newblock {\em Pattern recognition 43}, 3 (2010), 706--719.

\bibitem{haralick1979statistical}
{\sc Haralick, R.~M.}
\newblock Statistical and structural approaches to texture.
\newblock {\em Proceedings of the IEEE 67}, 5 (1979), 786--804.

\bibitem{haralick1973}
{\sc Haralick, R.~M., Shanmugam, K., and Dinstein, I.~H.}
\newblock Textural features for image classification.
\newblock {\em IEEE Transactions on systems, man, and cybernetics}, 6 (1973),
  610--621.

\bibitem{huang2006extreme}
{\sc Huang, G.-B., Zhu, Q.-Y., and Siew, C.-K.}
\newblock Extreme learning machine: theory and applications.
\newblock {\em Neurocomputing 70}, 1 (2006), 489--501.

\bibitem{idrissa2002texture}
{\sc Idrissa, M., and Acheroy, M.}
\newblock Texture classification using gabor filters.
\newblock {\em Pattern Recognition Letters 23}, 9 (2002), 1095--1102.

\bibitem{1990gabor}
{\sc Jain, A.~K., and Farrokhnia, F.}
\newblock Unsupervised texture segmentation using {G}abor filters.
\newblock In {\em Systems, Man and Cybernetics, 1990. Conference Proceedings.,
  IEEE International Conference on\/} (1990), IEEE, pp.~14--19.

\bibitem{kannala2012bsif}
{\sc Kannala, J., and Rahtu, E.}
\newblock Bsif: Binarized statistical image features.
\newblock In {\em Pattern Recognition (ICPR), 2012 21st International
  Conference on\/} (2012), IEEE, pp.~1363--1366.

\bibitem{kim1999statistical}
{\sc Kim, J.~K., and Park, H.~W.}
\newblock Statistical textural features for detection of microcalcifications in
  digitized mammograms.
\newblock {\em IEEE transactions on medical imaging 18}, 3 (1999), 231--238.

\bibitem{lam1997rotated}
{\sc Lam, W.-K., and Li, C.-K.}
\newblock Rotated texture classification by improved iterative morphological
  decomposition.
\newblock {\em IEE Proceedings-Vision, Image and Signal Processing 144}, 3
  (1997), 171--179.

\bibitem{rLEH51a}
{\sc Lehmer, D.~H.}
\newblock Mathematical methods in large scale computing units.
\newblock {\em Annals Comp. Laboratory Harvard University 26\/} (1951),
  141--146.

\bibitem{liu2018texturesurvey}
{\sc Liu, L., Chen, J., Fieguth, P., Zhao, G., Chellappa, R., and Pietikainen,
  M.}
\newblock A survey of recent advances in texture representation.
\newblock {\em arXiv preprint arXiv:1801.10324\/} (2018).

\bibitem{lfd}
{\sc Maani, R., Kalra, S., and Yang, Y.-H.}
\newblock Noise robust rotation invariant features for texture classification.
\newblock {\em Pattern Recognition 46}, 8 (2013), 2103--2116.

\bibitem{industrial2003}
{\sc Malamas, E.~N., Petrakis, E.~G., Zervakis, M., Petit, L., and Legat,
  J.-D.}
\newblock A survey on industrial vision systems, applications and tools.
\newblock {\em Image and vision computing 21}, 2 (2003), 171--188.

\bibitem{manjunath1996texture}
{\sc Manjunath, B.~S., and Ma, W.-Y.}
\newblock Texture features for browsing and retrieval of image data.
\newblock {\em IEEE Transactions on pattern analysis and machine intelligence
  18}, 8 (1996), 837--842.

\bibitem{Miranda2016}
{\sc Miranda, G. H.~B., Machicao, J., and Bruno, O.~M.}
\newblock Exploring spatio-temporal dynamics of cellular automata for pattern
  recognition in networks.
\newblock {\em Scientific Reports 6\/} (Nov 2016), 37329.

\bibitem{Moore1920}
{\sc Moore, E.~H.}
\newblock On the reciprocal of the general algebraic matrix.
\newblock {\em Bulletin of the American Mathematical Society 26\/} (1920),
  394--395.

\bibitem{nanni2012survey}
{\sc Nanni, L., Lumini, A., and Brahnam, S.}
\newblock Survey on {LBP} based texture descriptors for image classification.
\newblock {\em Expert Systems with Applications 39}, 3 (2012), 3634--3641.

\bibitem{OjalaMPVKH02}
{\sc Ojala, T., M{\"a}enp{\"a}{\"a}, T., Pietik{\"a}inen, M., Viertola, J.,
  Kyll{\"o}nen, J., and Huovinen, S.}
\newblock Outex: New framework for empirical evaluation of texture analysis
  algorithms.
\newblock In {\em International Conference on Pattern Recognition\/} (2002),
  pp.~701--706.

\bibitem{ojala2002multiresolution}
{\sc Ojala, T., Pietikainen, M., and Maenpaa, T.}
\newblock Multiresolution gray-scale and rotation invariant texture
  classification with local binary patterns.
\newblock {\em IEEE Transactions on pattern analysis and machine intelligence
  24}, 7 (2002), 971--987.

\bibitem{lpq}
{\sc Ojansivu, V., and Heikkil{\"a}, J.}
\newblock Blur insensitive texture classification using local phase
  quantization.
\newblock In {\em International conference on image and signal processing\/}
  (2008), Springer, pp.~236--243.

\bibitem{PALM2004integrativeCooccurrence}
{\sc Palm, C.}
\newblock Color texture classification by integrative co-occurrence matrices.
\newblock {\em Pattern recognition 37}, 5 (2004), 965--976.

\bibitem{panjwani1995markov}
{\sc Panjwani, D.~K., and Healey, G.}
\newblock Markov random field models for unsupervised segmentation of textured
  color images.
\newblock {\em IEEE Transactions on pattern analysis and machine intelligence
  17}, 10 (1995), 939--954.

\bibitem{pao1994learning}
{\sc Pao, Y.-H., Park, G.-H., and Sobajic, D.~J.}
\newblock Learning and generalization characteristics of the random vector
  functional-link net.
\newblock {\em Neurocomputing 6}, 2 (1994), 163--180.

\bibitem{pao1992functional}
{\sc Pao, Y.-H., and Takefuji, Y.}
\newblock Functional-link net computing: theory, system architecture, and
  functionalities.
\newblock {\em Computer 25}, 5 (1992), 76--79.

\bibitem{park1988random}
{\sc Park, S.~K., and Miller, K.~W.}
\newblock Random number generators: good ones are hard to find.
\newblock {\em Communications of the ACM 31}, 10 (1988), 1192--1201.

\bibitem{penrose_1955}
{\sc Penrose, R.}
\newblock A generalized inverse for matrices.
\newblock {\em Mathematical Proceedings of the Cambridge Philosophical Society
  51}, 3 (1955), 406–--413.

\bibitem{Vistex1995}
{\sc Picard, R., Graczyk, C., Mann, S., Wachman, J., Picard, L., and Campbell,
  L.}
\newblock {\em Vision texture database}.
\newblock Media Laboratory, MIT, Cambridge, Massachusetts, 1995.

\bibitem{Ribas2015}
{\sc Ribas, L.~C., Gon{\c{c}}alves, D.~N., Oru{\^{e}}, J. P.~M., and
  Gon{\c{c}}alves, W.~N.}
\newblock {Fractal dimension of maximum response filters applied to texture
  analysis}.
\newblock {\em Pattern Recognition Letters 65\/} (2015), 116--123.

\bibitem{scabini2019multilayer}
{\sc Scabini, L.~F., Condori, R.~H., Gonçalves, W.~N., and Bruno, O.~M.}
\newblock Multilayer complex network descriptors for color-texture
  characterization.
\newblock {\em Information Sciences 491\/} (2019), 30 -- 47.

\bibitem{scabini2015texture}
{\sc Scabini, L.~F., Gon{\c{c}}alves, W.~N., and Castro~Jr, A.~A.}
\newblock Texture analysis by bag-of-visual-words of complex networks.
\newblock In {\em Iberoamerican Congress on Pattern Recognition\/} (2015),
  Springer International Publishing, pp.~485--492.

\bibitem{schmidt1992feedforward}
{\sc Schmidt, W.~F., Kraaijveld, M.~A., and Duin, R. P.~W.}
\newblock Feedforward neural networks with random weights.
\newblock In {\em Proceedings., 11th IAPR International Conference on Pattern
  Recognition. Vol.II. Conference B: Pattern Recognition Methodology and
  Systems\/} (1992), pp.~1--4.

\bibitem{tiknonov1963}
{\sc Tikhonov, A.~N.}
\newblock On the solution of ill-posed problems and the method of
  regularization.
\newblock {\em Dokl. Akad. Nauk USSR 151}, 3 (1963), 501–--504.

\bibitem{smallworldCN}
{\sc Watts, D.~J., and Strogatz, S.~H.}
\newblock Collective dynamics of ‘small-world’networks.
\newblock {\em nature 393}, 6684 (1998), 440--442.

\bibitem{2013geology}
{\sc Wenk, H.~R.}
\newblock {\em Preferred Orientation in Deformed Metal and Rocks: An
  Introduction to Modern Texture Analysis}.
\newblock Elsevier, 2013.

\bibitem{weszka1976comparative}
{\sc Weszka, J.~S., Dyer, C.~R., and Rosenfeld, A.}
\newblock A comparative study of texture measures for terrain classification.
\newblock {\em IEEE transactions on Systems, Man, and Cybernetics}, 4 (1976),
  269--285.

\bibitem{xu2015complex}
{\sc Xu, D., Chen, X., Xie, Y., Yang, C., and Gui, W.}
\newblock Complex networks-based texture extraction and classification method
  for mineral flotation froth images.
\newblock {\em Minerals Engineering 83\/} (2015), 105--116.

\bibitem{zhang2002survey}
{\sc Zhang, J., and Tan, T.}
\newblock Brief review of invariant texture analysis methods.
\newblock {\em Pattern recognition 35}, 3 (2002), 735--747.

\bibitem{methods}
{\sc Zhang, J., and Tan, T.}
\newblock Brief review of invariant texture analysis methods.
\newblock {\em Pattern recognition 35}, 3 (2002), 735--747.

\bibitem{zhu2015adaptive}
{\sc Zhu, Z., You, X., Chen, C.~P., Tao, D., Ou, W., Jiang, X., and Zou, J.}
\newblock An adaptive hybrid pattern for noise-robust texture analysis.
\newblock {\em Pattern Recognition 48}, 8 (2015), 2592--2608.

\bibitem{zimer2011investigation}
{\sc Zimer, A.~M., Rios, E.~C., Mendes, P. d. C.~D., Gon{\c{c}}alves, W.~N.,
  Bruno, O.~M., Pereira, E.~C., and Mascaro, L.~H.}
\newblock Investigation of aisi 1040 steel corrosion in h2s solution containing
  chloride ions by digital image processing coupled with electrochemical
  techniques.
\newblock {\em Corrosion Science 53}, 10 (2011), 3193--3201.

\end{thebibliography}


\end{document}